\documentclass{article}

\usepackage{microtype}
\usepackage{graphicx}
\usepackage{subfigure}
\usepackage{booktabs} 

\usepackage{hyperref}

\usepackage[accepted]{icml2025}

\newcommand{\dockerhublink}{\url{https://hub.docker.com/r/gaunab/bencher}}
\newcommand{\bencherscaffoldlink}{\url{https://pypi.org/project/bencherscaffold/}}
\newcommand{\bencherclientlink}{\url{https://github.com/LeoIV/bencherclient}}
\newcommand{\bencherlink}{\url{https://github.com/LeoIV/bencher}}

\usepackage{amsmath}
\usepackage{amssymb}
\usepackage{mathtools}
\usepackage{amsthm}

\usepackage[capitalize,noabbrev]{cleveref}

\usepackage{bm}
\usepackage{xspace}
\usepackage{listings}
\usepackage{tabularx}
\usepackage{dirtree}
\usepackage{enumitem}
\usepackage{MnSymbol}
\usepackage{acronym}

\acrodef{RPC}{remote procedure call}
\acrodef{HPC}{high-performance computing}

\DeclareMathOperator*{\argmax}{arg\,max}

\theoremstyle{plain}

\theoremstyle{definition}

\theoremstyle{remark}

\newcommand{\method}{\texttt{Bencher}\xspace}
\usepackage[textsize=tiny]{todonotes}

\newcommand{\mytitle}{\texttt{Bencher}: Simple and Reproducible Benchmarking for Black-Box Optimization}

\icmltitlerunning{\mytitle}

\begin{document}

    \twocolumn[
        \icmltitle{\mytitle}



        \icmlsetsymbol{equal}{*}

        \begin{icmlauthorlist}
            \icmlauthor{Leonard Papenmeier}{lund}
            \icmlauthor{Luigi Nardi}{lund,dbtune}
        \end{icmlauthorlist}

        \icmlaffiliation{lund}{Department of Computer Science, Lund University, Sweden}
        \icmlaffiliation{dbtune}{DBtune}

        \icmlcorrespondingauthor{Leonard Papenmeier}{leonard.papenmeier@cs.lth.se}

        \icmlkeywords{Machine Learning, ICML}

        \vskip 0.3in
    ]



\printAffiliationsAndNotice{}  

\begin{abstract}
    We present \method, a modular benchmarking framework for black-box optimization that fundamentally decouples benchmark execution from optimization logic.
    Unlike prior suites that focus on combining many benchmarks in a single project, \method introduces a clean abstraction boundary: each benchmark is isolated in its own virtual Python environment and accessed via a unified, version-agnostic \ac{RPC} interface.
    This design eliminates dependency conflicts and simplifies the integration of diverse, real-world benchmarks, which often have complex and conflicting software requirements.
    \method can be deployed locally or remotely via Docker or on \ac{HPC} clusters via Singularity, providing a containerized, reproducible runtime for any benchmark.
    Its lightweight client requires minimal setup and supports drop-in evaluation of 80 benchmarks across continuous, categorical, and binary domains.
\end{abstract}

\acresetall
\section{Introduction}\label{sec:intro}

Black-box optimization refers to the problem of optimizing a function $\bm{x}^{*} = \argmax_{\bm{x} \in \mathcal{X}} f(\bm{x})$ of unknown form for which we can only observe the function value but no derivatives~\citep{turner2021bayesian}.
In particular, $f$ may be highly multimodal and/or noisy.
Furthermore, $f$ often is expensive to evaluate, and we can only afford to spend a limited number of function evaluations to find a good solution.
Problems of this type have received considerably interest due to their ubiquity in fields like chemical engineering~\citep{lobato2017parallel,burger2020mobile}, engineering~\citep{lam2018advances,maathuis2024high}, hyperparameter optimization~\citep{snoek2012practical,NIPS2011_86e8f7ab}, and life sciences~\citep{tallorin2018discovering,cosenza2022multi}.

The development of sample-efficient algorithms for black-box optimization is a highly active research field.
To give a holistic picture of an algorithm's performance, it is usually evaluated on a wide range of benchmarks.
Besides synthetic benchmarks of known form, methods are usually evaluated on various benchmarks reflecting real-world applications.
We call these benchmarks \emph{real-world benchmarks}.
Running these benchmarks often requires significant effort for two reasons.
First, many benchmarks have very specific software requirements and can have a complex setup, making it hard to set them up even in a greenfield environment.
For example, the Mujoco benchmarks used in~\citet{wang2020learning} require several scientific libraries to be installed, setting environment variables, and the presence of the Mujoco executable in a specific location.
Second, benchmark code can be outdated, requiring specific versions of \texttt{Python} and other external packages.
For instance, the benchmarks used by~\citet{wang2018batched} require Python version 3.8, which has reached end-of-life in 2024, conflicting with newer Python versions.
This can lead to dependency conflicts when running multiple benchmarks in the same environment or when the code of the algorithm to be evaluated requires newer versions of \texttt{Python} or other packages.

Setting up benchmarks is often a non-trivial task and not the main focus of a research project.
This can lead to cases where results on the same benchmark are not comparable, arguably due to inconsistent setups.
For example, the \texttt{TuRBO} baseline~\cite{eriksson2019scalable} in \citet[Fig. 3, Hopper benchmark]{fan2024minimizing} lies on a completely different scale than the same baseline on the same benchmark in \citet[Fig. 3, Hopper benchmark]{nguyen2022local}.

This paper introduces \method\footnote{\bencherlink}, a benchmarking framework that allows running benchmarks in a reproducible and simple way.
\method isolates every benchmark (or set of compatible benchmarks) in a virtual \texttt{Python} environment.
This allows each benchmark to run with different versions of \texttt{Python} and other external packages.
To facilitate the setup of a benchmark, \method can run in a Docker container.
We provide a simple package with minimal external dependencies that is responsible for the communication with the \method Docker container.
Furthermore, we provide a Singularity container that can be used to run \method on a cluster and provide detailed instructions for doing so.
This setup decouples the benchmarking code from the optimization algorithm, allowing for more freedom in the dependency specification of the optimizer's code.
Furthermore, it allows for easy integration of new benchmarks, as they can be added as new subprojects in the \method repository, using their own Python environment and dependencies.

In summary, we make the following contributions
\begin{itemize}[leftmargin=*,topsep=0pt]
    \setlength\itemsep{0em}
    \item We introduce \method, a benchmarking framework that ensures reproducibility and simplicity by isolating each benchmark in a dedicated virtual Python environment.
    \item We design a server-based architecture that handles \acp{RPC} and enables flexible, version-independent execution of benchmarks.
    \item We provide containerized solutions (Docker and Singularity) along with lightweight client packages to support easy deployment on both local machines and \ac{HPC} clusters.
\end{itemize}

\section{Related Work}\label{sec:related_work}
Several benchmark suites for non-convex black-box optimization have been proposed in the literature.
A recent survey by \citet{sala2020benchmarking} provides a comprehensive overview of the state of the art in black-box optimization benchmarks.

Nevergrad~\citep{bennet2021nevergrad} is an open-source platform for black-box optimization that offers a wide portfolio of algorithms and benchmark problems, including synthetic, combinatorial, and real-world tasks.
It features automatic algorithm selection, extensive parallelism, and a public leaderboard, but does not support environment isolation, limiting the simultaneous use of conflicting benchmarks.
Many of the benchmarks in Nevergrad are undocumented, as acknowledged by the authors\footnote{See \url{https://facebookresearch.github.io/nevergrad/benchmarks.html\#list-of-benchmarks}}.
For this reason, Nevergrad is currently not included in \method but we aim to establish a collaboration with the authors to add the benchmarks in the future.

COCO~\citep{nikolaus_hansen_2019_2594848} is a long-standing and widely used benchmarking platform for zero-order black-box optimization, including noisy, multi-objective, and non-continuous problems.
The single-objective, continuous, and noise-free benchmarks from the BBOB suite are especially popular and have been used in many studies.
These methods are included in \method, using the implementation provided by \citet{de2024iohexperimenter}.

IOHexperimenter~\citet{de2024iohexperimenter} is the experimentation module of IOHprofiler~\citet{doerr2018iohprofiler}, offering a wide range of continuous and pseudo-boolean benchmarks.
\method implements most of the benchmarks from IOHexperimenter, including the BBOB, pseudo-boolean, W-model, and submodular benchmarks.

While all of the aforementioned benchmark suites advance the state of the art in black-box optimization, they differ from \method in that they do not aim at decoupling the benchmarks but instead focus on providing a comprehensive set of benchmarks in a single package.

\begin{figure}
    \centering
    \includegraphics[width=.9\linewidth]{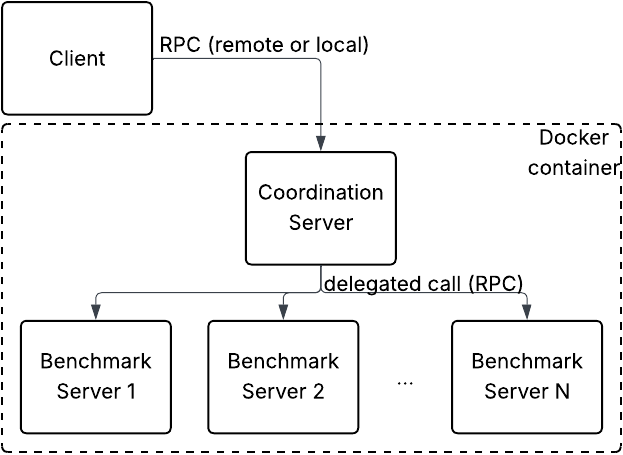}
    \caption{
        The \method architecture.
        The server runs in a Docker container and listens to \acp{RPC} from clients, which can be on the same or on a different machine.
        The server is composed of multiple \texttt{Poetry} environments, one for each benchmark.
    }
    \label{fig:bencher_scheme}
\end{figure}

\section{\method: Design and Implementation}\label{sec:design}
\method follows a client-server architecture (see \cref{fig:bencher_scheme}).
The server is responsible for running the benchmarks and listening to \acp{RPC} from clients.
It can be run in a Docker container, which allows running the server in an isolated environment with minimal setup.
Furthermore, a Singularity container can inherit the Docker container to run \method on an \ac{HPC} cluster.

The server is responsible for running the benchmarks and listening to \acp{RPC} from clients.
It is composed of multiple \texttt{Poetry} environments -- one coordinator and multiple environments for sets of compatible benchmarks.
The coordinator is responsible for listening to \acp{RPC} and delegating them to the appropriate benchmark environment.
The client and the server, and the different benchmark environments among themselves, communicate via \texttt{gRPC}\footnote{\url{https://grpc.io/}}, a high-performance, open-source universal RPC framework.
Each benchmark environment exposes a \texttt{gRPC} service on a specific port that listens for incoming \acp{RPC}.
The clients exclusively communicate with the coordinator, which forwards the \acp{RPC} to the appropriate benchmark environment.
This structure has the advantage that the Docker container only needs to expose a single port and that the clients only need to speak to the coordinator.

\subsection{Client}\label{sec:client}
The client is the interface to the server and the main entry point for users.
It establishes a connection to the coordination server (see \cref{fig:bencher_scheme}) and provides methods to evaluate points on the benchmarks.
The client is implemented in \texttt{Python} and is available on PyPI (\bencherscaffoldlink).
It is designed to have minimal external dependencies (only \texttt{grpcio} and \texttt{protobuf} are required) and can be installed by \texttt{pip install bencherscaffold}.
Once the client successfully connects to the server, it can be used to evaluate points on the benchmarks as follows:

\begin{lstlisting}[breaklines=true,prebreak=\mbox{\ensuremath{\rhookswarrow}},caption=Exemplary client code for evaluating a benchmark (long lines are broken).]
from bencherscaffold.client import BencherClient
from bencherscaffold.protoclasses.bencher_pb2 import Value, ValueType
client = BencherClient()
benchmark_name = 'mopta08'
values = [Value(type=ValueType.CONTINUOUS, value=0.5) for _ in range(124)]
result: float = client.evaluate_point(benchmark_name, point)
\end{lstlisting}

This code runs the 124-dimensional \texttt{Mopta08} vehicle mass optimization benchmark~\citep{jones2008large} in its soft-constrained version~\citep{eriksson2019scalable}.
All 124 parameters of this benchmark are continuous and normalized to the unit hypercube $[0,1]^{124}$.
An exemplary implementation running all available benchmarks is available on GitHub (\bencherclientlink).
This repository is also used during the testing of \method to ensure that all benchmarks are working correctly.

\subsection{Docker Container}
The server can be run in a Docker container, which allows running the server in an isolated environment with minimal setup.
The Docker container can be built from source by cloning the repository and running \texttt{docker build} in the root directory of the repository.
It is also available on Docker Hub (\dockerhublink).
Since all communication between the different benchmark environments happens internally, the Docker container only needs to expose a single port to the outside world.
The container can be pulled and run in the background with the following command:
\begin{lstlisting}[language=bash,breaklines=true,postbreak=\mbox{$\hookrightarrow$\space}]
docker pull gaunab/bencher
docker run -d --name bencher -p 50051:50051 gaunab/bencher
\end{lstlisting}

\subsection{HPC Setup}
\method can be run on an \ac{HPC} cluster using Singularity with minimal manual setup by inheriting the Docker container.
One constraint is that the Singularity container needs to be started as an instance in the background.
To isolate instances running on the same cluster node, we recommend using a unique instance name for each \ac{HPC} job.
With \texttt{slurm}, this can be done by using the job ID as the instance name, e.g.,
\begin{lstlisting}[language=bash,breaklines=true,postbreak=\mbox{$\hookrightarrow$\space}]
singularity instance start `INST_NAME'
\end{lstlisting}
where `\texttt{INST\_NAME}' is replaced with \texttt{\$\{SLURM\_ARRAY\_JOB\_ID\}\_\$\{SLURM\_ARRAY\_TASK\_ID\}}.
A job can then be run with
\begin{lstlisting}[language=bash,breaklines=true,prebreak=\mbox{\ensuremath{\rhookswarrow}}]
singularity run instance://INST_NAME CMD
\end{lstlisting}

\subsection{Project Structure}
The \method implementation follows a prespecified structure that allows for easy extendability.
\begin{figure}
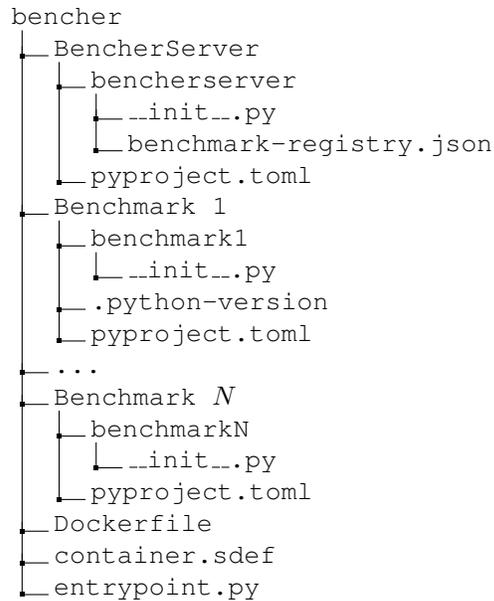

\dirtree{%
.1 bencher.
.2 BencherServer.
.3 bencherserver.
.4 \_\_init\_\_.py.
.4 benchmark-registry.json.
.3 pyproject.toml.
.2 Benchmark 1.
.3 benchmark1.
.4 \_\_init\_\_.py.
.3 .python-version.
.3 pyproject.toml.
.2 \ldots.
.2 Benchmark $N$.
.3 benchmarkN.
.4 \_\_init\_\_.py.
.3 pyproject.toml.
.2 Dockerfile.
.2 container.sdef.
.2 entrypoint.py.
}
\caption{The \method directory structure.}
\label{fig:bencher_structure}
\end{figure}
The directory structure is shown in \cref{fig:bencher_structure}.
All benchmarks, the server, and other relevant files are located in the \texttt{bencher} directory.
We refer to each of these subdirectories as \emph{subprojects}.
Each subproject has its own \texttt{pyproject.toml} file, which defines the dependencies for the subproject and a starting script \texttt{start-benchmark-service}.
The \texttt{entrypoint.py}, which is the entry point for the Docker container, goes into each subproject, activates the virtual environment, and starts the server using this script.

The \texttt{benchmark-registry.json} file is used to define the mapping between the benchmark names and the ports on which the benchmarks are running:
\begin{lstlisting}[language=bash,breaklines=true,postbreak=\mbox{$\hookrightarrow$\space}]
... "lasso-dna": {
    "port": 50053,
    "dimensions": 180,
    "type": "purely_continuous"
}, ...
\end{lstlisting}
where \texttt{lasso-dna} is the name of the 180-dimensional \texttt{Lasso-DNA} benchmark~\citep{vsehic2022lassobench}, \texttt{port} is the port on which the benchmark is running, \texttt{dimensions} is the number of dimensions of the benchmark, and \texttt{type} is the type of the benchmark.
For instance, \texttt{purely\_continuous} means that all dimensions of the benchmark are continuous and \texttt{binary} means that all dimensions of the benchmark are binary.

Some subprojects may have the \texttt{.python-version} file.
This file is used during the Docker installation of the project to set the correct \texttt{Python} version for the subproject.
Different Python versions are managed by \texttt{pyenv}\footnote{\url{https://github.com/pyenv/pyenv}} and are installed in the Docker container.
Subprojects without this file will use the default \texttt{Python} version of the Docker container.

When using Docker, the \texttt{entrypoint.py} file is executed when the container is started.
When using Singularity, the \texttt{entrypoint.py} file \emph{must} be defined as the startscript container definition file:
\begin{lstlisting}[language=bash,breaklines=true,postbreak=\mbox{$\hookrightarrow$\space}]
%startscript
bash -c "python3.11 /entrypoint.py"
\end{lstlisting}
An exemplary container definition file is provided in the repository.
New benchmarks can be added by creating a new subproject with the same structure as the existing ones.
In particular, the project must define a \texttt{pyproject.toml} file, a \texttt{start-benchmark-service} script, and, optionally, a \texttt{.python-version} file.

\subsection{Testing}

\method is tested using integration testing.
Triggered by a commit or a daily cron job, the full Docker container is built and started in the background.
Then, the \texttt{bencherclient} (see~\cref{sec:client}) is used to run all benchmarks.
If all benchmarks are running successfully, the Docker container is pushed to Docker Hub, and the Singularity container is built, inheriting from the previously pushed Docker image.

\section{Benchmarks}
\method currently supports 80 benchmarks, including 18 real-world benchmarks.
We list all benchmarks in \cref{sec:benchmarks} and restrict this section to a general overview.
\method currently is limited to unconstrained, single-objective optimization problems.
Benchmarks can be continuous, ordinal, binary, or categorical.
All continuous benchmarks are normalized to the unit hypercube $[0,1]^d$.
Categorical benchmarks, such as \texttt{pestcontrol}, expect an integer input for each dimension; the number of categories is documented in the \texttt{README.md} file in the repository.

Most of the benchmarks implemented in \method are well-known benchmarks from the literature.
For instance, the soft-constrained version of the \texttt{Mopta08} benchmark~\citep{jones2008large} was originally introduced in \citet{eriksson2019scalable} but has found widespread adoption in the high-dimensional Bayesian optimization literature~\citep{shen2021computationally,eriksson2021high,papenmeier2022increasing,papenmeier2023bounce,hvarfner2024vanilla,xu2024standard,papenmeier2025understanding}.
However, the original link to the executables of the benchmark is dead, and they are currently only available since other researchers uploaded them, as acknowledged by \citet{xu2024standard}.
Similarly, the Mujoco benchmarks used in various papers, including \citet{wang2020learning,papenmeier2022increasing,hvarfner2024vanilla}, require installing additional software, setting environment variables, and downloading additional executables, potentially reducing adoption and reproducibility.

\section{Conclusion and Future Work}\label{sec:conclusion}

We present \method, a benchmarking framework for black-box optimization that allows running benchmarks in a reproducible and simple way.
It largely decouples the benchmarking code from the optimization algorithm, allowing for more freedom in the dependency specification of the optimizer's code.
\method's client only requires minimal dependencies to communicate with the server and is easily installable via \texttt{pip}.
The server can be run in a Docker container and abstracts away the complexity of setting up the benchmarks for the user.
The benchmarks are isolated in their own virtual environments, allowing for different versions of \texttt{Python} and other external packages.
\method further provides a Singularity container that can be used to run \method on a cluster and provides detailed instructions for doing so.

We plan to extend \method in several ways.
Currently, \method supports unconstrained, single-objective optimization problems.
While this covers a wide range of applications, we aim for a more general framework that can also handle constrained, multi-fidelity, and multi-objective optimization problems.
In the future, we will also support more diverse search domains, covering, for instance, graph-based benchmarks.
Furthermore, we plan to gradually extend the set of benchmarks in \method, adding benchmark suites like the one in Nevergrad~\citep{bennet2021nevergrad}, \texttt{CATBench}~\cite{torring2024catbench}, and \texttt{HPOBench}~\citep{eggensperger2021hpobench} suite.


\newpage

\bibliography{bibliography}
\bibliographystyle{icml2025}

\newpage
\appendix
\onecolumn

\section{Benchmark Overview}\label{sec:benchmarks}

\begin{table}[H]
\centering
\begin{tabularx}{\linewidth}{lX}
\toprule
Source & Benchmarks \\
\midrule
\citet{vsehic2021lassobench} & \texttt{lasso-simple} ($d=60$, cont.), \texttt{lasso-medium} ($d=100$, cont.), \texttt{lasso-high} ($d=300$, cont.), \texttt{lasso-hard} ($d=1000$, cont.),  \texttt{lasso-breastcancer} ($d=10$, cont.), \texttt{lasso-diabetes} ($d=8$, cont.), \texttt{lasso-leukemia} ($d=7129$, cont.), \texttt{lasso-dna} ($d=180$, cont.), \texttt{lasso-rcv1} ($d=19959$, cont.)\\
\citet{eriksson2019scalable} & \texttt{mopta08} ($d=124$, cont., see also \citet{jones2008large}), \texttt{rover} ($d=60$, cont.), \texttt{robotpushing} ($d=14$, cont., see also \citet{wang2018batched}), \texttt{lunarlander} ($d=12$, cont.) \\
\citet{deshwal2023bayesian}  & \texttt{maxsat60} ($d=60$, binary)\\
\citet{papenmeier2023bounce} & \texttt{maxsat125} ($d=125$, binary)\\
\citet{eriksson2021scalable} & \texttt{svm} ($d=388$, cont., this is an adapted version introduced in \citet{papenmeier2022increasing})\\
\citet{wang2020learning}     & \texttt{mujoco-ant} ($d=888$, cont.), \texttt{mujoco-hopper} ($d=33$, cont.), \texttt{mujoco-walker} ($d=102$, cont.), \texttt{mujoco-halfcheetah} ($d=102$, cont.), \texttt{mujoco-swimmer} ($d=16$, cont.), \texttt{mujoco-humanoid} ($d=6392$, cont.)\\
\citet{oh2019combinatorial}  & \texttt{pestcontrol} ($d=25$, cat.)\\
\citet{nikolaus_hansen_2019_2594848,doerr2018iohprofiler} & \texttt{bbob-sphere} (cont.), \texttt{bbob-ellipsoid} (cont.), \texttt{bbob-rastrigin} (cont.), \texttt{bbob-buecherastrigin} (cont.), \texttt{bbob-linearslope} (cont.), \texttt{bbob-attractivesector} (cont.), \texttt{bbob-stepellipsoid} (cont.), \texttt{bbob-rosenbrock} (cont.), \texttt{bbob-rosenbrockrotated} (cont.), \texttt{bbob-ellipsoidrotated} (cont.), \texttt{bbob-discus} (cont.), \texttt{bbob-bentcigar} (cont.), \texttt{bbob-sharpridge} (cont.), \texttt{bbob-differentpowers} (cont.), \texttt{bbob-rastriginrotated} (cont.), \texttt{bbob-weierstrass} (cont.), \texttt{bbob-schaffers10} (cont.), \texttt{bbob-schaffers1000} (cont.), \texttt{bbob-griewankrosenbrock} (cont.), \texttt{bbob-schwefel} (cont.), \texttt{bbob-gallagher101} (cont.), \texttt{bbob-gallagher21} (cont.), \texttt{bbob-katsuura} (cont.), \texttt{bbob-lunacekbirastrigin} (cont.)\\
\citet{doerr2018iohprofiler} & \texttt{pbo-onemax} (binary), \texttt{pbo-leadingones} (binary), \texttt{pbo-linear} (binary), \texttt{pbo-onemaxdummy1} (binary), \texttt{pbo-onemaxdummy2} (binary), \texttt{pbo-onemaxneutrality} (binary), \texttt{pbo-onemaxepistasis} (binary), \texttt{pbo-onemaxruggedness1} (binary), \texttt{pbo-onemaxruggedness2} (binary), \texttt{pbo-onemaxruggedness3} (binary), \texttt{pbo-leadingonesdummy1} (binary), \texttt{pbo-leadingonesdummy2} (binary), \texttt{pbo-leadingonesneutrality} (binary), \texttt{pbo-leadingonesepistasis} (binary), \texttt{pbo-leadingonesruggedness1} (binary), \texttt{pbo-leadingonesruggedness2} (binary), \texttt{pbo-leadingonesruggedness3} (binary), \texttt{pbo-labs} (binary), \texttt{pbo-isingring} (binary), \texttt{pbo-isingtorus} (binary), \texttt{pbo-isingtriangular} (binary), \texttt{pbo-mis} (binary), \texttt{pbo-nqueens} (binary), \texttt{pbo-concatenatedtrap} (binary), \texttt{pbo-nklandscapes} (binary)\\
\bottomrule
\end{tabularx}
\end{table}

\end{document}